\documentclass{ecai} 

\usepackage{latexsym}
\usepackage{amssymb}
\usepackage{amsmath}
\usepackage{amsthm}
\usepackage{booktabs}
\usepackage{enumitem}
\usepackage{graphicx}
\usepackage{color}
\usepackage{multirow}

\newcommand{\BibTeX}{B\kern-.05em{\sc i\kern-.025em b}\kern-.08em\TeX}

\begin{document}


\begin{frontmatter}

\paperid{1947} 

\title{Attribute Guidance With Inherent Pseudo-label For Occluded Person Re-identification}

\author[A]{\fnms{Rui}~\snm{Zhi}}
\author[A]{\fnms{Zhen}~\snm{Yang}}
\author[A]{\fnms{Haiyang}~\snm{Zhang}\thanks{Corresponding Author. Email: zhhy@bupt.edu.cn}} 

\address[A]{Beijing University of Post and Telecommunication} 

\begin{abstract}
Person re-identification (Re-ID) aims to match person images across different camera views, with occluded Re-ID addressing scenarios where pedestrians are partially visible. While pre-trained vision-language models have shown effectiveness in Re-ID tasks, they face significant challenges in occluded scenarios by focusing on holistic image semantics while neglecting fine-grained attribute information. This limitation becomes particularly evident when dealing with partially occluded pedestrians or when distinguishing between individuals with subtle appearance differences.

To address this limitation, we propose Attribute-Guide ReID (AG-ReID), a novel framework that leverages pre-trained models' inherent capabilities to extract fine-grained semantic attributes without additional data or annotations. Our framework operates through a two-stage process: first generating attribute pseudo-labels that capture subtle visual characteristics, then introducing a dual-guidance mechanism that combines holistic and fine-grained attribute information to enhance image feature extraction.

Extensive experiments demonstrate that AG-ReID achieves state-of-the-art results on multiple widely-used Re-ID datasets, showing significant improvements in handling occlusions and subtle attribute differences while maintaining competitive performance on standard Re-ID scenarios.
\end{abstract}

\end{frontmatter}


\section{Introduction}

\begin{figure}[t]
    \centering
    \includegraphics[width=\linewidth]{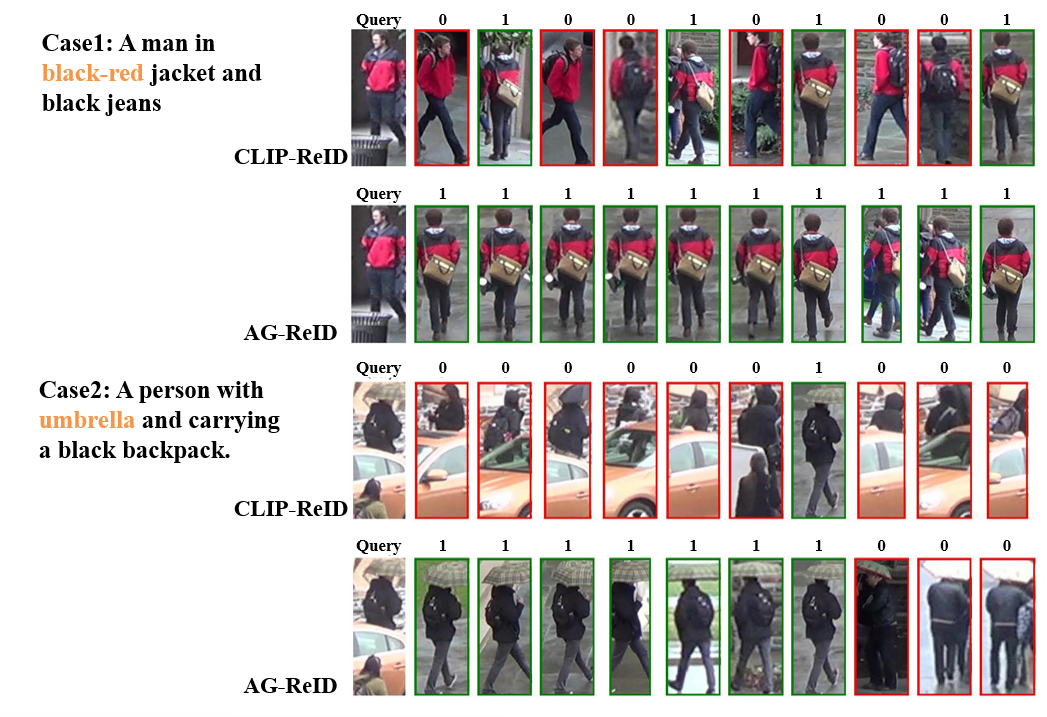}
    \caption{Comparison of retrieval results between CLIP-ReID and AG-ReID on challenging cases. Case1: A person wearing a black-red jacket with subtle color variations. Case2: A person with an umbrella in occluded scenario. For each case, the first column shows the query image, followed by top-10 retrieval results. Incorrect matches (marked as '0') are outlined in red, while correct matches (marked as '1') are outlined in green.}
    \label{fig:result}
\end{figure}

Person Re-Identification (ReID) is a computer vision task that aims to match person images across different camera views \citep{reid_survey}. With the rapid development of smart cities and surveillance systems, ReID has become increasingly crucial in various real-world applications, processing millions of surveillance videos daily. It requires the extraction of unique features from images to recognize the same person, even when there are alterations in pose, lighting, and viewpoint. Among various challenges in ReID, occlusion presents a particularly significant obstacle. This necessitates advanced algorithms capable of effectively managing partial occlusions while maintaining precise identification capabilities.

Existing approaches to address occlusion in ReID can be categorized based on their feature extraction strategies. Global methods, which are commonly used in general ReID tasks, learn a single feature to represent the entire image. While effective in standard scenarios, these methods often fail to handle occluded regions effectively as they cannot distinguish between visible and occluded parts. To address this limitation, researchers have proposed part-based techniques \citep{horeid,pat,bpbreid} and attribute-based methods. Part-based techniques focus on comparing visible body parts between images, which helps to identify individuals even when some parts are occluded. Recent studies \citep{luperson,mpreid,mals} have shown that attribute-based methods can significantly enhance model performance by leveraging detailed visual characteristics such as clothing style, accessories, and physical features. These two approaches have shown promising results in handling occluded scenarios, but they inevitably require additional supervision signals, such as pose estimation data, human parsing labels, or manually annotated attributes. This dependence on extra annotation data significantly limits their practical application in large-scale scenarios.

Recent advances in pre-trained vision-language models present a promising direction for addressing these limitations. Studies \citep{clipreid,mpreid} have established that models like CLIP (Contrastive Language-Image Pre-training) \citep{clip} are highly effective in extracting comprehensive semantic information from images, achieving significant performance improvements in ReID tasks. By learning combined representations of text and visuals through contrastive learning on massive image-text pairs, these models excel at grasping complex semantic concepts and transferring knowledge to downstream tasks. However, in the context of occlusion ReID tasks, these models face similar challenges as global methods. Relying solely on holistic information from textual prompts to guide image feature extraction has significant limitations. This approach leads to the omission of fine-grained semantic information within the image, causing the pre-trained model to focus only on primary features that may be partially visible or occluded. As a result, performance degrades substantially in occlusion scenarios. As illustrated in Figure \ref{fig:result}, the tendency of CLIP-ReID to focus on holistic features while overlooking fine-grained details significantly impacts its ability to distinguish between individuals with similar appearances or identify partially occluded persons.

This paper explores the potential of mining fine-grained feature semantic information from the pre-trained vision-language model CLIP to improve its performance in occluded ReID tasks. Our approach leverages attribute guidance, thereby eliminating the necessity for additional data or capabilities and representing a promising avenue for enhancing the model's efficacy in this challenging domain. We propose a novel attribute-guided method called AG-ReID that supports this challenging task through a two-stage training process. In the first stage, AG-ReID acquires image attribute pseudo-labels by leveraging the detailed semantic information within CLIP through context optimization. In the second stage, it guides the extraction of image features with both holistic and fine-grained attribute information, improving the retrieval performance of the model. Specifically, the dual guidance includes: 1) Attribute-prompt guidance: using the overall attribute prompt text feature to guide image features through contrastive learning. 2) Fine-grained attribute pseudo-label guidance: the learnable tokens are implicitly trained through the CoOp \citep{coop} method to obtain fine-grained semantics, thereby guiding the extraction of image features. Furthermore, we propose an attribute encoder for aligning image features with attribute pseudo-labels, and an attribute loss for measuring the semantic difference between them. To handle inconsistent features in occluded scenarios, we introduce a noise-masking mechanism that selectively considers attribute pairs based on their semantic similarity. The efficacy of AG-ReID was assessed through experimentation on multiple well-known occluded and holistic ReID datasets. The results demonstrated that AG-ReID outperformed a number of existing methods. The primary contributions of this work are summarized as follows:

\begin{itemize}
    \item[$\bullet$] To the best of our knowledge, this is the first attempt to improve occluded ReID by embedding fine-grained attribute semantic into image features through the inherent capabilities of the CLIP model, without the need for extra data or annotations, significantly enhancing model performance in occlusion scenarios.
    \item[$\bullet$] We present a novel attribute dual-guidance ReID framework called AG-ReID, which effectively guides the extraction of image features through both holistic text features and fine-grained attribute pseudo-labels, improving feature accuracy and robustness.
    \item[$\bullet$] We propose an innovative method that implicitly trains attribute pseudo-labels through context optimization, and design corresponding attribute encoder module and attribute loss to achieve effective alignment between image features and attribute pseudo-labels.
    \item[$\bullet$] We conduct evaluations on multiple challenging datasets, including Occluded-ReID, P-Duke and MSMT17, demonstrating that AG-ReID achieves state-of-the-art performance in both occluded and holistic scenarios.
\end{itemize}

\section{Related Works}

\subsection{Pre-trained Vision-Language Learning}
Pre-trained vision-language models are a class of machine learning models trained on large-scale datasets to understand and process both visual and textual data. These models can understand the relationships between images and text, enabling them to perform a variety of tasks such as image captioning, visual question answering, and image-text retrieval.
The power of these models lies in their ability to generalize well to various downstream tasks. This is achieved by pre-training on extensive datasets, which enables the models to learn a wide range of visual-textual relationships. Once pre-trained, these models can be fine-tuned on specific tasks, making them versatile and effective in real-world applications.

The dual encoder architecture is a prevalent pre-trained vision-language model architecture that employs two separate unimodal encoders to independently process images and text. It utilizes shallow attention layers or dot products to align the embeddings of both modalities into a unified semantic space, enhancing efficiency in tasks like image-text retrieval. Nonetheless, the limited depth of interaction between the modalities can pose challenges in complex visual-language understanding tasks. Both the widely used CLIP and ALIGN \citep{align} models incorporate this dual encoder architecture.

Contrastive Language–Image Pre-training (CLIP) is a pre-trained model introduced by OpenAI that efficiently learns visual concepts from natural language supervision and is adaptable to various downstream tasks \citep{clip}. It consists of two encoders, an Image Encoder and a Text Encoder. Image encoders, with architectures like ViT \citep{vit} or ResNet50 \citep{resnet}, are designed to transform images from a high-dimensional RGB space into a low-dimensional embedding space. The text encoder converts each word in the given prompt to a unique numeric ID, maps IDs into embedding vectors, and finally encodes them to a text feature that contains prompt semantic information. During the training phase, CLIP optimizes a symmetric cross entropy loss to achieve the target of maximizing the cosine similarity for matched pairs while minimizing the cosine similarity for all other unmatched pairs.

One of the challenges in applying pre-trained models to downstream tasks is the time-consuming and domain-expertise-required prompt engineering. Context Optimization (CoOp) automates this process by modeling prompt context words using learnable vectors while keeping the pre-trained parameters frozen \citep{coop}. CoOp significantly enhances prompt engineering performance and demonstrates robust domain generalization capabilities compared to manual prompts. In essence, CoOp transforms static text prompts into learnable text templates. It acquires and learns text descriptions directly through the intrinsic multi-modal abilities of the pre-trained model, avoiding intricate manual word tuning. Specifically, the prompt embedding given to the text encoder is designed with trainable tokens and fixed tokens, where the number and position of trainable tokens can be adjusted according to the requirements of downstream tasks.

\subsection{Attribute-based Person ReID}
Person ReID has been extensively studied due to its critical applications in surveillance and security. Traditional person ReID methods primarily focus on visual features extracted from images, and combine additional information such as pose, body mask, and visible infrared to address occlusion, light changes, and more \citep{horeid, song2018mask, bpbreid, infrared}. Similarly, recent research has explored the use of fine-grained attributes, such as clothing color, accessories, and physical characteristics, to enhance ReID performance. 

Attribute-based Person ReID leverages these descriptive attributes to provide additional semantic information, improving the robustness and accuracy of person matching. Notable work in this area includes artificially annotating image attribute features in existing datasets \citep{li2019attribute}, using language models to automatically generate and utilize attribute descriptions \citep{mals}, and integrating multi-modal data to bridge the gap between textual and visual information \citep{mpreid}. Recent advances have also explored prompt-guided approaches for feature disentangling in occluded scenarios \citep{cui2024profd} and text-based multi-granularity contrastive learning for occluded person ReID \citep{wu2024text}.

In addition, recent work has also explored the interpretability of attribute-guided methods. For example, AMD method \citep{chen2021explainable} provides post-mortem explanations for existing ReID models by identifying and quantifying the contributions of different attributes. 

\section{Method}

\begin{figure*}[t]
    \centering
    \includegraphics[width=\textwidth]{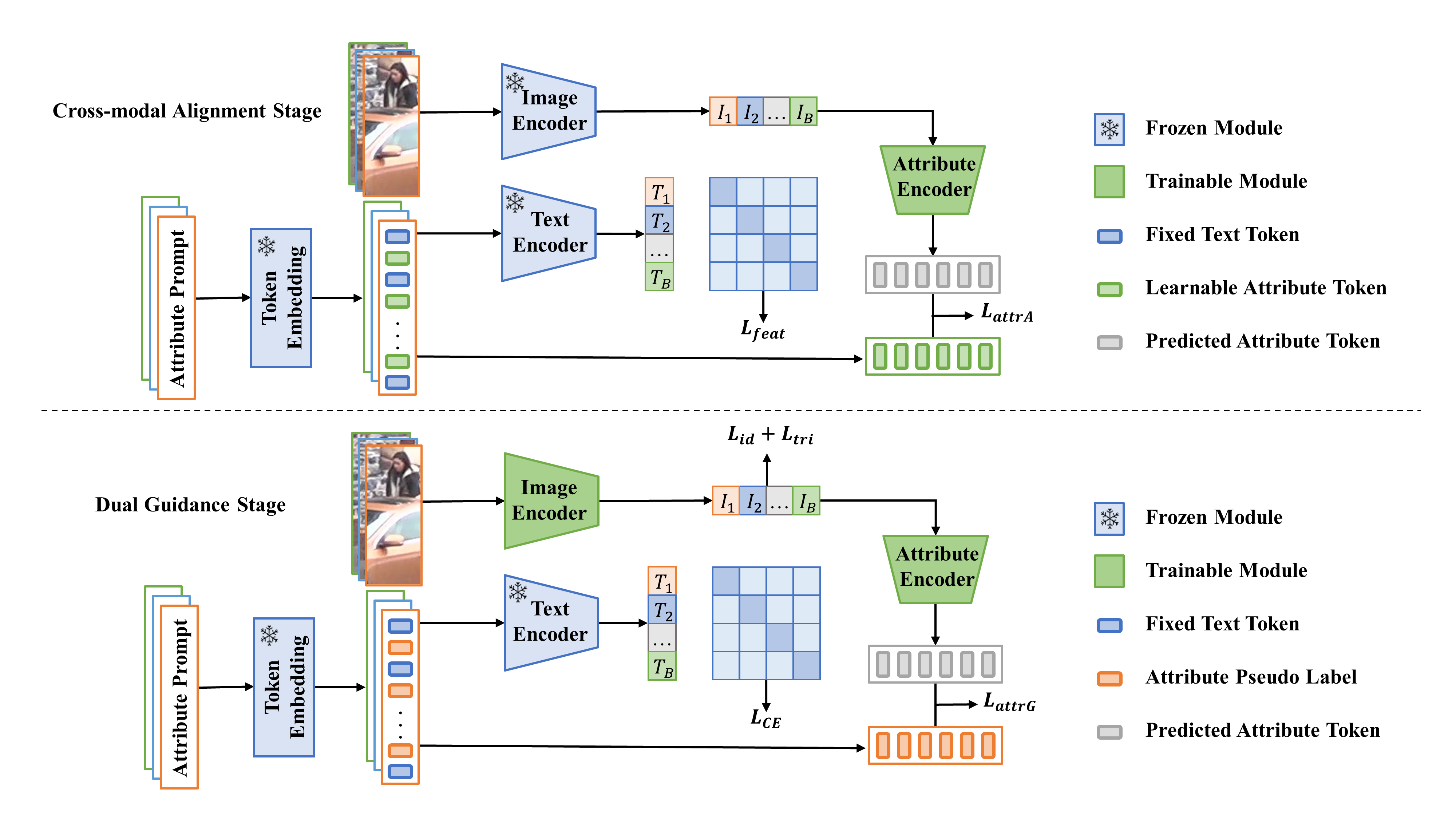}
    \caption{Overview of the AG-ReID framework's two-stage training. 
    \textbf{Stage 1 (Cross-modal Alignment):} Using frozen Image/Text Encoders, learnable tokens within attribute prompts are trained via contrastive loss (\(L_{\text{feat}}\)) to generate attribute pseudo-labels. Simultaneously, a trainable Attribute Encoder learns to predict attribute tokens aligned with these pseudo-labels via \(L_{\text{attrA}}\). 
    \textbf{Stage 2 (Dual Guidance):} The Image Encoder is trained with guidance from both attribute-prompt text features (\(L_{\text{CE}}\)) and the generated attribute pseudo-labels (\(L_{\text{attrG}}\)), leveraging a fine-tuned Attribute Encoder. Standard ReID losses (\(L_{\text{id}}, L_{\text{tri}}\)) are also applied to the image features. The legend indicates module status (frozen/trainable) and token types.}
    \label{fig:arch}
\end{figure*}

\subsection{Overview}
This section provides a detailed introduction to the AG-ReID model, which leverages the rich semantic information inherent in pre-trained models to enhance image retrieval performance for occluded person ReID tasks. Our framework operates through a two-stage process that aligns visual and textual modalities. In the first stage, we establish a semantic bridge between images and their fine-grained attributes, generating attribute pseudo-labels that capture subtle visual characteristics. In the second stage, we introduce a dual-guidance mechanism that combines holistic attribute-prompt features and fine-grained attribute pseudo-labels to enhance image feature extraction. To handle inconsistent features in occluded scenarios, we propose a noise-masking mechanism that selectively focuses on reliable attribute matches while filtering out those affected by occlusions.

\subsection{Cross-Modal Alignment Stage}
\subsubsection{Preliminary}
Our approach is built upon a dual-encoder architecture designed to process visual and textual information. Specifically, $i \in \{1...B\}$ denotes the index within a batch. Given the image batch $M = \{ m_1, m2, ..., m_B \}$ and their text descriptions $P = \{ p_1, p2, ..., p_B \}$, an image encoder $E_M$ maps images to 512-d features $f^M_i = E_{M}(m_i)$. For text prompts $p_i$, each prompt is first tokenized (e.g., using byte pair encoding) and then embedded into 512-d word tokens $T_i$ via an embedding layer $\text{Embd}$. These tokens are subsequently processed by a text encoder $E_T$ to produce the final text feature $f^T_i$.

\begin{equation}
T_i = [t]^i_1 [t]^i_2 ... [t]^i_k = \text{Embd}(p_i)
\end{equation}

\begin{equation}
f^T_i = E_T(T_i)
\end{equation}

During pre-training, such architectures typically optimize a symmetric contrastive loss to align the image and text representations in a shared embedding space. This objective maximizes the cosine similarity for matched image-text pairs while minimizing it for unmatched pairs:

\begin{equation}
s(f_M, f_T) = \text{L2}(\text{proj}_M(f_M)) \cdot \text{L2}(\text{proj}_T(f_T))^T
\end{equation}

where $\text{proj}_M$ and $\text{proj}_T$ are projection layers, $\text{L2}$ denotes L2-normalization, and $t$ is a learned temperature parameter.

To adapt the text encoder for specific downstream tasks without extensive prompt engineering, we leverage a technique that incorporates learnable context vectors into the prompt embedding. Specifically, the input $T_i$ to the text encoder $E_T$ is structured to include $r$ trainable vectors $[v]$ alongside $k-r$ fixed tokens $[t]$. The number $r$ and the position of $[v]$ can be adjusted based on task requirements:

\begin{equation}
T_i = [t]^i_1 [v]^i_1 ... [v]^i_r [t]^i_{k-r}
\label{eq: coop_token}
\end{equation}

During fine-tuning on a downstream task, only these learnable vectors $[v]$ are optimized, utilizing the semantic knowledge encoded within the frozen pre-trained model parameters to capture task-relevant information.

\subsubsection{Attribute Prompt Template}
To effectively capture fine-grained attributes of person images, we design a set of attribute prompt templates that cover various visual characteristics. These templates are constructed using a combination of fixed and learnable tokens, following the CoOp framework. The attribute prompt templates are designed to describe different aspects of a person's appearance, including:

\begin{itemize}
    \item Clothing attributes (e.g., color, style, pattern)
    \item Accessories (e.g., bags, hats, glasses)
    \item Body characteristics (e.g., height, build)
    \item Pose and movement
\end{itemize}

Each attribute prompt template follows the structure:

\begin{equation}
T_{\text{attr}} = [t]_{\text{prefix}} [v]_1 [v]_2 ... [v]_r [t]_{\text{suffix}}
\end{equation}

where $[t]_{\text{prefix}}$ and $[t]_{\text{suffix}}$ are fixed tokens that provide context, and $[v]_1$ through $[v]_r$ are learnable tokens that adapt to capture specific attribute information. The number of learnable tokens $r$ is determined based on the complexity of the attribute being described.

\subsubsection{Attribute Pseudo-label Generation}
Inspired by the positive impact of fine-grained attributes in ReID, we utilize the trainable attribute prompt structure defined above. Unlike holistic prompts used in some prior work \citep{clipreid}, our attribute prompts feature learnable tokens dispersed throughout the text, each intended to capture a specific fine-grained feature. This design is critical for recognizing subtle differences, especially under occlusion, setting the stage for effective pseudo-label generation as described next.

In our framework, attribute pseudo-labels refer to learnable token representations automatically derived from CLIP and CoOp, requiring no manual annotation. These pseudo-labels capture fine-grained visible attribute features (e.g., hair style, clothing color, accessories) and serve as auxiliary supervision signals to guide the image encoder towards learning more discriminative attribute-aware features. Unlike traditional labels that require human annotation, our pseudo-labels are semantically trained to describe the collective attribute characteristics of all images belonging to the same identity.

Our selection of attributes is informed by prior work demonstrating their effectiveness in person ReID. Specifically, we adopt key attribute categories identified as discriminative in ATPM \citep{mals}, encompassing aspects like gender, hairstyle, posture, and clothing characteristics. Furthermore, the designed prompt template module offers inherent flexibility; it facilitates the straightforward incorporation of alternative attribute sets or descriptive contexts, such as explicitly mentioning potential occlusion, through configuration adjustments without requiring modifications to the underlying code structure.

Given a ReID dataset batch of $n$ images $M = \{m_1, m_2, ..., m_n\}$, representing $K$ unique identities, let $P = \{p_1, p_2, ..., p_K\}$ be the corresponding attribute prompts for each identity. For an image $m_i$ associated with identity $k$, its corresponding attribute prompt $p_k$ contains $r$ learnable tokens, as defined previously. We denote this set of learnable tokens as $V_i = \{[v]^k_1, [v]^k_2, ..., [v]^k_r\}$.

In the alignment stage of training, the text features $f_T$ of attribute prompt are aligned with the image features $f_M$ by SupConLoss \citep{supconloss}, which is improved on top of the cross-entropy loss function for the supervised contrastive learning task.

\begin{equation}
L_{\text{feat}} = \text{SupCon}(f_M, f_T) + \text{SupCon}(f_T, f_M)
\end{equation}

In this way, not only the text feature $f_T$ gets the semantics corresponding to the image, but also each learnable token $[v]$ gets the semantics of its corresponding attribute. At this time, we will call the learnable attribute token set $V$ as semantics attribute pseudo-label.

To use attribute pseudo-labels for guiding image features during the following stage, we propose the attribute encoder $E_A$ with 4 self-attention layers to align image feature $f^i_M$ with attribute pseudo-labels $V_i$, which predicts attributes set $A_i$ by image features.

\begin{equation}
A_i = [a^i_1, a^i_2, ..., a^i_r] = E_A(f^i_M)
\end{equation}

The attribute align loss $L_{\text{attrA}}$ in this stage is formulated as,

\begin{equation}
L_{\text{attrA}} = - \frac{1}{r} \sum_{i=1}^{n} \sum_{j=1}^{r} \cos(a^i_j, v^i_j)
\end{equation}

The loss function in the cross-modal alignment stage $L_{\text{align}}$ consists of $L_{\text{feat}}$ and $L_{\text{attrA}}$, and $\lambda$ is a hyper-parameter for adjusting loss weights.

\begin{equation}
L_{\text{align}} = L_{\text{feat}} + \lambda L_{\text{attrA}}
\end{equation}

\subsection{Dual Guidance Stage}
Building on the cross-modal alignment stage, we obtain text features encapsulating the holistic semantic information of the text prompt, the attribute pseudo-label denoting fine-grained attribute information, and an attribute encoder designed to align image features with their respective attribute features.
At this stage, the integration of both holistic and fine-grained information guides the feature extraction process for images, which is why it is referred to as the dual guidance stage.

The former computes the loss using the cross entropy loss function $L_{\text{CE}}$, which is frequently employed in contrastive learning, whereas the latter is optimized by the improved previously mentioned attributes alignment loss $L_{\text{attrG}}$.

\begin{equation}
L_{\text{CE}} = \text{CrossEntropy}(f_M, f_T)
\end{equation}

Additionally, the ID loss $L_{\text{id}}$ and triplet loss $L_{\text{tri}}$, commonly employed in ReID tasks, are also combined.

\begin{equation}
L_{\text{id}} = - \sum_{i=1}^{n} q_i \log(p_i)
\end{equation}

\begin{equation}
L_{\text{tri}} = \max(d_p - d_n + \alpha,0)
\end{equation}

where $q_i$ denotes the value in the target distribution, and $p_i$ represents ID prediction logits of class $i$, $d_p$ and $d_n$ are feature distances of positive pair and negative pair, while $\alpha$ is the margin of triple loss.

The idea of fine-tuning image feature encoder directly using pseudo-labels is expected to yield good results in holistic datasets, and this is confirmed in subsequent experiments.

However, it is crucial to recognize that images with the same ID do not always share consistent semantics. In other words, attribute pseudo-labels reflect the collective attribute semantics of all images with a common ID, which may cause some tokens $a$ predicted by the attribute encoder to mismatch their pseudo-labels $v$ in the occluded case. Such mismatched pairs are referred to as ``noise''. This issue arises when parts of some images differ from the majority within the same ID group. For instance, if the lower half of a person is visible in most images with ID $i$, but obscured in others, the corresponding pseudo-label $v$ will predominantly contain the semantic information of the visible parts. This leads to the token $a$ at the corresponding positions in the occluded image having semantic discrepancies, potentially misleading the feature extraction process.

To address this issue, we suggest a slight modification to the attribute alignment loss for occluded ReID datasets, termed the noise-mask method. In this approach, only pairs with a similarity above the threshold $\gamma$ will be considered for loss calculation, while the rest will be masked. For holistic datasets, $\gamma$ is assigned a value of $-1$, which means that loss will be computed for every pair.

\begin{equation}
D(i,j) =
\begin{cases}
1   & \text{if}  \cos(a^i_j, t^i_j) > \gamma \\
0   & \text{otherwise.} \\
\end{cases}
\end{equation}

Thus, the attribute pseudo-label guidance loss $L_{\text{attrG}}$ is calculated as follows,

\begin{equation}
L_{\text{attrG}} = - \frac{1}{r} \sum_{i=1}^{n} \sum_{j=1}^{r}D(i,j) \cdot \cos(a^i_j, v^i_j)
\end{equation}

The final dual guidance stage loss is formulated as,

\begin{equation}
L_{\text{guide}} = L_{\text{id}} + L_{\text{tri}} + L_{\text{CE}} + \beta L_{\text{attrG}}
\end{equation}

which consists of three components: the basic ReID losses $L_{\text{id}}$ and $L_{\text{tri}}$, the text prompt guidance loss $L_{\text{CE}}$, and the attribute pseudo-label guidance loss $L_{\text{attrG}}$. These components work together to guide the effective extraction of image features, where $\beta$ is the weight of $L_{\text{attrG}}$.

\section{Experiments}

\subsection{Experimental Settings}
\subsubsection{Datasets and Evaluation Protocols}

\begin{table}[htbp]
    \begin{center}
    \begin{tabular}{c|ccc}
         \hline
         Dataset & Image & ID & Camera \\
         \hline
         Occ-Duke & 35,489 & 1,404 & 8 \\
         Occ-ReID & 2000 & 200 & 6 \\
         P-Duke & 24143 & 1299 & 2 \\
         \hline
         MSMT17 & 126,441 & 4,101 & 15 \\
         Market-1501 & 32,668 & 1,501 & 6 \\
         DukeMTMC & 36,411 & 1,404 & 8 \\
         \hline
    \end{tabular}
    \caption{Datasets Statistics.}
    \label{tab:datasets_info}
    \end{center}
\end{table}

We evaluated the proposed AG-ReID method on both occluded, partial, and holistic ReID benchmarks, including Occluded-Duke \cite{occduke}. Occluded-ReID \cite{occreid_survey}, P-DukeMTMC \cite{occreid_survey}, MSMT17 \cite{msmt17}, Market-1501 \cite{market1501_map} and DukeMTMC-reID \cite{duke}. The details of the datasets are summarized in Tab. \ref{tab:datasets_info}. 

Following common practices, we employ the cumulative matching characteristics (CMC) \cite{cmc} at Rank-1 (R1) and the mean average precision (mAP) \cite{market1501_map} for performance evaluation.

\subsubsection{Implementation Details}
Our method is implemented on a server equipped with a NVIDIA GeForce RTX 3090 Ti GPU. We utilize the pre-trained CLIP with ViT-B/16 as the backbone, resizing all images to 256 $\times$ 128. During the training phase, data augmentation techniques such as random horizontal flipping, cropping, and erasing \cite{random_earsing} are used. In the cross-modal alignment stage, the Adam optimizer is used with a base learning rate of 3.5e-04 and a warm-up learning rate that increases linearly from 1e-06. In the dual guidance stage, the base learning rate is set at 5e-06. The batch size is set to 32, with K to 4 instances sampled for each identity per batch. In addition, side information embeddings (SIE) and overlap patches (OLP) are utilized to further improve the model performance \cite{transreid}. The margin for triplet loss $\alpha$ is set to 0.3, and the weight $\lambda$ and $\beta$ in the loss function are set to 1 and 0.01, respectively.

During training, the attribute encoder introduces approximately 2.76M additional parameters, which represents only about 3.2\% of the ViT-B/16 backbone (86M parameters). Crucially, during inference, AG-ReID only requires the image encoder (ViT-B/16), resulting in identical computational complexity and model size as CLIP-ReID and other ViT-based methods. The attribute encoder is exclusively used during training and does not affect deployment efficiency.

The selection of specific attributes in the prompt-attribute templates refers to the table of attributes available in ATPM \cite{mals}. For threshold $\gamma$ in the dual guidance stage, we choose the Otsu method \cite{otsu} for occluded datasets to automatically calculate the threshold to maximize the variance between classes.

\subsection{Performance Comparison}
\subsubsection{Performance on Occluded datasets}

\begin{table*}[htbp]
    \begin{center}
    \begin{tabular}{c|c|c|c|cc|cc|cc}
         \hline
         \multirow{2}{*}{Method} & \multirow{2}{*}{Reference} & \multirow{2}{*}{Category} & \multirow{2}{*}{Detail} & \multicolumn{2}{c|}{Occ-Duke} & \multicolumn{2}{c|}{Occ-ReID} & \multicolumn{2}{c}{P-Duke}\\
         & & & & mAP & R@1 & mAP & R@1 & mAP & R@1 \\
         \hline
         HOReID & CVPR2020 & \multirow{6}{*}{Extra data} & Pose Estimator & 43.8 & 55.1 & 70.2 & 80.3 & - & - \\
         PAT & CVPR2021 & & Part Prototype & 53.6 & 64.5 & 72.1 & 81.6 & - & 88.0\\
         FED & CVPR2022 & & Manually Crop & 56.4 & 68.1 & 79.3 & 86.3 & 80.5 & 83.1 \\
         PFD & AAAI2022 & & Pose Estimator & 61.8 & 69.5 & 83.0 & 81.5 & - & -\\
         BPBReID$_{RI}$ & WACV2023 & & Human parsing labels & 57.5 & 71.3 & 70.9 & 77.0 & 79.2 & 91.3\\
         KRP$_{IN}$ & ECCV2024 & & Manual keypoint labels & \textbf{67.1} & \textbf{79.8} & 79.1 & 85.4 & - & 86.0\\
         \hline
         DPM-SPT & AAAI2024 &  \multirow{2}{*}{Pretrain model} & Pre-trained ViT & 63.0 & 74.7 & 81.1 & 87.8 & - & -\\
         ADP & AAAI2024 & & Pre-trained ViT & 63.8 & 74.5 & 82.0 & 88.2 & - & -\\
         \hline
         CLIP-ReID & AAAI2023 & Baseline & Pre-trained CLIP & 60.0 & 67.8 & - & - & - & -\\
         \hline
         AG-ReID & - & Our & Pre-trained CLIP & 63.2 & 70.4 & \textbf{87.6} & \textbf{90.1} & \textbf{84.5} & \textbf{91.8} \\
         \hline
    \end{tabular}
    \caption{Performance on occluded datasets.}
    \label{tab:occ_result}
    \end{center}
\end{table*}

Table \ref{tab:occ_result} presents the comparison between AG-ReID and state-of-the-art methods, including recent approaches like KRP \cite{somers2024keypoint} and ADP \cite{adp}, on three widely-used occluded/partial person ReID datasets: Occluded-Duke, Occluded-ReID, and P-Duke. Compared to the baseline CLIP-ReID on the Occluded-Duke dataset, our method achieves a 3.2\% improvement in mAP and a 2.6\% increase in R@1. Besides, AG-ReID achieves state-of-the-art mAP and R@1 performance on both Occluded-ReID and P-Duke datasets. This highlights the effectiveness of fully utilizing the fine-grained semantic information within the pre-trained model without requiring extra annotation data. Furthermore, applying standard k-reciprocal re-ranking significantly boosts the performance on these challenging datasets (e.g., reaching 75.4\% mAP and 74.1 R@1 on Occ-Duke). Detailed re-ranking results are provided in supplementary materials.

We primarily report results using ViT-B/16 as it generally outperforms ResNet-50 in recent ReID works. For completeness, we also evaluated AG-ReID with ResNet-50 backbone on Occluded-ReID, achieving 55.1\% mAP and 63.7\% Rank-1, compared to 53.5\% mAP and 61.0\% Rank-1 for CLIP-ReID (RN50), demonstrating the effectiveness of our approach across different architectures.

Case studies in Figure \ref{fig:result} further reveal that our method effectively addresses errors caused by overlooking fine-grained differences, aligning with our initial design expectations. 

\subsubsection{Performance on Holistic datasets}

\begin{table}[htbp]
    \begin{center}
    \begin{tabular}{c|cc|cc|cc}
         \hline
         \multirow{2}{*}{Method} & \multicolumn{2}{c|}{MSMT17}  & \multicolumn{2}{c|}{Market-1501} & \multicolumn{2}{c}{DukeMTMC} \\
         & mAP & R@1 & mAP & R@1 & mAP & R@1 \\
         \hline
         PAT & - & - & 88.0 & 95.4 & 78.2 & 88.8\\
         FED & - & - & 86.3 & 95.0 & 78.0 & 89.4 \\
         BPBReID$_{RI}$ & - & - & 88.4 & 95.7 & 81.3 & \textbf{91.7}\\
         DPM-SPT & - & - & 89.4 & 95.5 & 82.4 & 91.1\\
         ADP & - & - & 89.5 & 95.6 & 83.1 & 91.2\\
         PFD & - & - & 89.7 & 95.5 & 83.2 & 91.2\\
         HOReID & 50.4 & 74.4 & 84.9 & 94.2 & 75.6 & 86.9\\
         CC-ViT & 64.6 & 83.7 & 90.4 & 96.0 & 81.2 & 90.4\\
         SOLIDER & 67.4 & 85.9 & \textbf{91.6} & \textbf{96.1} & - & -\\
         TransReID & 67.4 & 85.3 & 88.9 & 95.3 & 82.0 & 90.7\\
         DiP & 71.8 & 87.3 & 90.3 & 95.8 & \textbf{85.2} & \textbf{91.7}\\
         \hline
         CLIP-ReID & 75.8 & 89.7 & 90.5 & 95.4 & 83.1 & 90.8 \\
         \hline
         AG-ReID & \textbf{76.7} & \textbf{90.1} & 90.8 & 95.8 & 83.3 & 91.0 \\
         \hline
    \end{tabular}
    \caption{Performance on holistic datasets.}
    \label{tab:holi_result}
    \end{center}
\end{table}

Table \ref{tab:holi_result} presents the results of AG-ReID compared to other methods on three widely used holistic person ReID datasets. Our model shows improvement over the CLIP-ReID baseline on all three datasets. Moreover, our approach attains state-of-the-art performance on the large-scale MSMT17 dataset, surpassing the compared methods. It is also observed that the CLIP-ReID baseline, while performing well on MSMT17, was less competitive on Market-1501 and DukeMTMC. This strong performance on MSMT17 likely benefits from the dataset's large scale and diversity, which may better align with the data distribution encountered during the model's extensive pre-training. When combined with k-reciprocal re-ranking, AG-ReID's performance is further enhanced on holistic datasets as well (e.g., achieving 86.6\% mAP and 91.1\% R@1 on MSMT17), demonstrating its compatibility with standard post-processing techniques.

\subsection{Ablation Studies and Analysis}

\subsubsection{Ablation study for dual guidance}

\begin{table}[htbp]
    \begin{center}
    \begin{tabular}{ccc|cccc}
         \hline
         \multicolumn{3}{c|}{Strategies} & \multicolumn{4}{c}{Result for Occ-Duke} \\
         Baseline & AT & AP & mAP & R@1 & R@5 & R@10 \\
         \hline
         \checkmark &  &  & 60.0 & 67.8 & 80.1 & 85.2 \\
         \checkmark & \checkmark &  & 62.4 & 69.7 & 81.7 & 86.2 \\
         \checkmark &  & \checkmark & 62.2 & 69.2 & 81.9 & 86.3 \\
         \checkmark & \checkmark & \checkmark & \textbf{63.2} & \textbf{70.4} & \textbf{82.4} & \textbf{86.4} \\
         \hline
    \end{tabular}
    \caption{Ablation studies for AG-ReID, where ``AT'' is attribute-prompt template guidance and ``AP'' is attribute pseudo-label guidance. The baseline is CLIP-ReID model with SIE and OLP.}
    \label{tab:ablation}
    \end{center}
\end{table}

To further confirm the significance of the attribute-prompt template (AT) and attribute pseudo-label guidance (AP) in our AG-ReID framework, we conduct ablation studies using the CLIP-ReID model with SIE and OLP as the baseline. As shown in Table \ref{tab:ablation}, introducing either the attribute prompt template or the attribute pseudo-label guidance individually yields performance improvements over the baseline. Combining both components (AG-ReID) further enhances the guiding effect of fine-grained attributes on image features, achieving the best results.

\subsubsection{Attribute Prompt Template}

\begin{table}[htbp]
    \begin{center}
    \begin{tabular}{c|cccc}
         \hline
         \multirow{2}{*}{Template} & \multicolumn{4}{c}{Occ-Duke} \\
         & mAP & R@1 & R@5 & R@10 \\
         \hline
         Default Prompt & 62.2 & 69.2 & 81.9 & 86.3 \\
         Longer Prompt & 62.0 & 69.9 & 82.1 & 86.2 \\
         Random Prompt & 62.1 & 69.5 & 81.6 & 86.1 \\
         Attribute Prompt a & 62.6 & 69.8 & 82.4 & 86.2 \\
         Attribute Prompt b & 63.0 & \textbf{70.5} & 82.0 & \textbf{86.5} \\
         Attribute Prompt c & 62.8 & 69.9 & \textbf{82.6} & 86.1 \\
         Attribute Prompt d & \textbf{63.2} & 70.4 & 82.4 & 86.4 \\
         \hline
    \end{tabular}
    \caption{Result comparison for different prompt template. The default prompt is the text prompt proposed by CLIP-ReID, the longer prompt increases the number of trainable tokens on the basis of default, while the random prompt makes the number of fixed tokens become 0, making all tokens learnable, and the attribute prompt a, b, c, d are our trainable attribute prompt templates with different attributes selection.}
    \label{tab:template}
    \end{center}
\end{table}

We investigate the impact of different prompt template designs in Table \ref{tab:template}. While our pseudo-label guidance mechanism can operate with various prompt types, we hypothesized that attribute-specific templates would better enrich the fine-grained semantics of the learned pseudo-labels compared to holistic prompts. The results confirm this hypothesis: templates explicitly structured around attributes (Attribute Prompt a, b, c, d) consistently outperform the default holistic prompt, longer holistic prompts, and random token prompts. This indicates that the core concept of attribute-structured prompting is beneficial, and suggests the method is relatively robust to specific template wording, reducing the need for extensive prompt engineering. The detailed content of these templates can be found in the supplementary materials.

\subsubsection{Dual guidance threshold}

\begin{table}[htbp]
    \begin{center}
    \begin{tabular}{c|cccc}
         \hline
         \multirow{2}{*}{Threshold} & \multicolumn{4}{c}{Occ-Duke} \\
         & mAP & R@1 & R@5 & R@10 \\
         \hline
         P50 & 61.9 & 69.9 & 82.7 & 86.0 \\
         P75 & 62.5 & \textbf{70.6} & \textbf{82.5} & 86.3 \\
         P90 & 62.8 & 70.5 & 82.3 & 86.0 \\
         Otsu & \textbf{63.2} & 70.4 & 82.4 & \textbf{86.4} \\
         \hline
    \end{tabular}
    \caption{Result comparison for different threshold $\gamma$.}
    \label{tab:threshold}
    \end{center}
\end{table}

In the dual guidance stage, to address the issue where samples within the same ID may exhibit inconsistent features in occluded datasets, we introduce the noise-mask method with an attribute loss threshold $\gamma$. Loss is calculated only for predicted attribute and pseudo-label pairs whose similarity exceeds this threshold. As shown in Table \ref{tab:threshold}, we compare different strategies for setting $\gamma$: fixed percentiles (P50, P75, P90) and the Otsu method \cite{otsu}, which automatically determines a threshold to maximize inter-class variance. 

Higher $\gamma$ values impose stricter constraints, ensuring only well-aligned attribute features contribute to the loss, mitigating noise from inconsistent image details. Experimental results demonstrate that the Otsu method effectively determines an optimal threshold autonomously, eliminating the need for manual tuning per dataset and achieving the best performance.

\subsubsection{Visualization}

\begin{figure}[t]
    \centering
    \includegraphics[width=\linewidth]{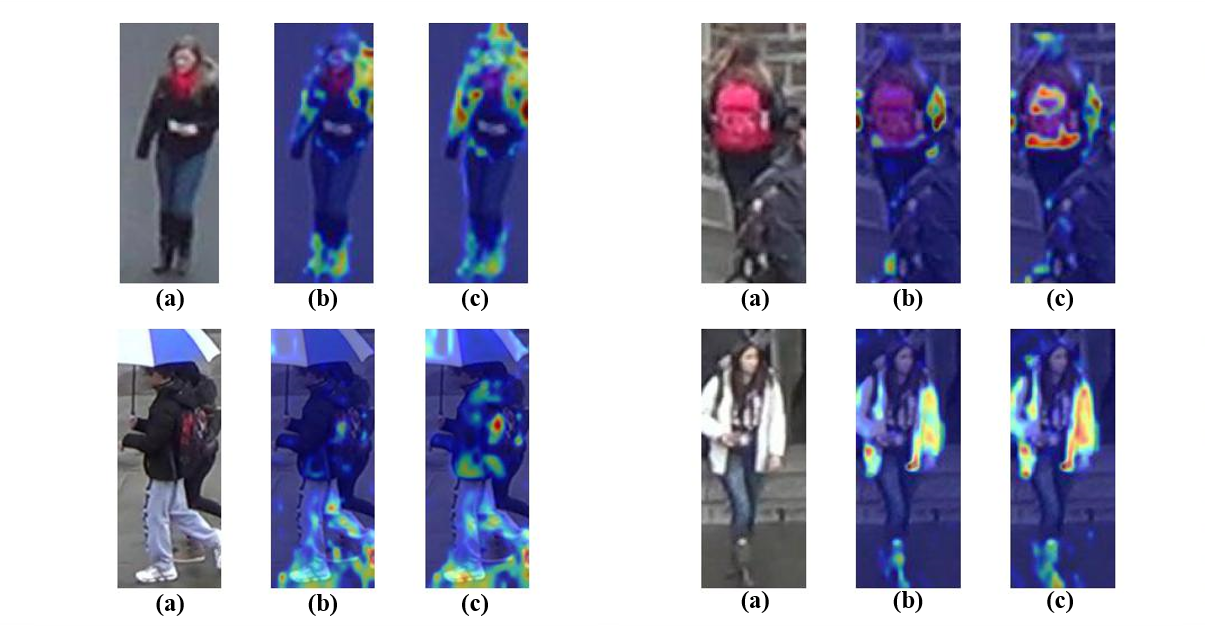}
    \caption{Heat map for visualization. (a) Source images, (b) Baseline, (d) AG-ReID.}
    \label{fig:heatmap}
    
\end{figure}

We employed the Grad-CAM method \cite{gradcam} to visualize the key regions of interest for the model, shown in Figure \ref{fig:heatmap}. Compared to the baseline model, our AG-ReID model exhibits a heightened focus on fine-grained attributes such as hair, bags, and clothing.

\subsubsection{Discussion}
Overall, our experiments validate the effectiveness of AG-ReID. The performance gains, particularly on occluded datasets, stem from the successful mining and utilization of fine-grained attribute semantics inherent within the pre-trained vision-language model. The dual-guidance mechanism ensures that both holistic context and detailed attributes contribute to the final feature representation, while the noise-masking strategy provides robustness against inconsistent visual cues common in occlusion. This approach, which avoids the need for external annotations or specialized modules like pose estimators, offers a promising direction for developing more practical and data-efficient ReID systems, potentially extendable to other fine-grained recognition tasks facing similar challenges of partial information or subtle distinctions.

\section{Conclusion}

In this paper, we introduce AG-ReID, a novel framework for occluded person re-identification that leverages the fine-grained semantics of pre-trained vision-language models. By employing attribute text prompts and attribute pseudo-labels, our method significantly enhances the descriptiveness of image features.

Through extensive experiments on multiple datasets, AG-ReID demonstrates superior performance compared to existing methods, achieving state-of-the-art results on both occluded and holistic ReID tasks. Specifically, our method shows significant improvements in handling challenging cases involving occlusions and subtle attribute differences, while maintaining competitive performance on standard ReID scenarios. The experimental results validate the effectiveness of our approach in mining and utilizing fine-grained attribute semantics inherent within pre-trained vision-language models.

This work highlights the potential of integrating fine-grained attribute information from pre-trained vision-language models to improve ReID accuracy without the need for additional data or complex manual annotations. Our approach offers several advantages: 1) It eliminates the dependency on external supervision signals such as pose estimation data or manually annotated attributes; 2) It effectively handles occluded scenarios by focusing on reliable attribute matches; 3) It maintains robust performance across different datasets and scenarios. Future work will explore the effectiveness of this approach in cross-dataset scenarios, as well as further optimization and application in real-world settings.




\bibliography{mybibfile}


\end{document}



\begin{frontmatter}

\paperid{1947} 

\title{Supplementary Material: Attribute Guidance With Inherent Pseudo-label For Occluded Person Re-identification}

\author[A]{\fnms{Rui}~\snm{Zhi}}
\author[A]{\fnms{Zhen}~\snm{Yang}}
\author[A]{\fnms{Haiyang}~\snm{Zhang}\thanks{Corresponding Author. Email: zhhy@bupt.edu.cn}}

\address[A]{Beijing University of Post and Telecommunication}

\end{frontmatter}

\section{Attribute encoder structure}
The attribute encoder consists of a multi-layer transformer architecture, which applies self-attention encoding to the image token list. The input sequence is derived from the token list produced by the image encoder.

Given that each image corresponds to $r$ attribute pseudo-labels, the attribute encoder can be implemented in two ways: 1) Create a distinct encoder for each attribute pseudo-label and train them separately, or 2) Train a single encoder for all pseudo-labels uniformly, and add trainable class tokens representing different attributes before outputting the token list. The latter can reduce the CUDA memory usage of the attribute encoder max to $\frac{1}{r}$, but only slightly affect performance.

The results are presented in Table \ref{tab:ae_implementation}, illustrating the impact of various attribute encoder implementations on the model performance. In this context, the subscript $SE$ denotes single encoder, while $ME$ represents multi encoder.

\begin{table}
    \begin{center}
    \begin{tabular}{c|cccc}
         \hline
         Implementation & mAP & R@1 & R@5 & R@10 \\
         \hline
         Occ-Duke$_{SE}$ & 62.8 & 70.1 & 82.3 & 86.2 \\
         Occ-Duke$_{ME}$ & 63.2 & 70.4 & 82.4 & 86.4 \\
         \hline
         Market-1501$_{SE}$ & 90.8 & 95.7 & 98.3 & 99.0 \\
         Market-1501$_{ME}$ & 90.8 & 95.8 & 98.6 & 99.0 \\
         \hline
    \end{tabular}
    \caption{Result for different attribute encoder implementations.}
    \label{tab:ae_implementation}
    \end{center}
\end{table}

\section{Re-ranking result}

Re-ranking is a widely recognized and effective post-processing method for the ReID task, which operates without the need for human interaction or labeled data. The primary objective of re-ranking is to enhance the final ranking by elevating the positions of relevant images through additional processing based on the initial ranking results.

One prevalent re-ranking technique is k-reciprocal encoding. The fundamental concept is that if a gallery image is similar to the probe query within the k-reciprocal nearest neighbors, it is more likely to be a true match.

Table \ref{tab:rerank_result} presents the results, with subscript $RK$ representing the results after re-ranking, demonstrating a significant improvement in the mAP indicators for most datasets following re-ranking. The anomalous performance observed in the Occ-ReID dataset may be attributed to its unique training methodology. Unlike other datasets, Occ-ReID lacks a direct training set and instead evaluates the model post pre-training on the Market-1501 dataset. This deviation in distribution between the test set and the training set likely contributes to the observed performance discrepancies after re-ranking.

\begin{table}
    \begin{center}
    \begin{tabular}{c|cccc}
         \hline
         Datasets & mAP & R@1 & R@5 & R@10 \\
         \hline
         Occ-Duke & 63.2 & 70.4 & 82.4 & 86.4 \\
         Occ-Duke$_{RK}$ & 75.4 & 74.1 & 82.3 & 85.2 \\
         Occ-ReID & 87.6 & 90.1 & 95.8 & 97.5 \\
         Occ-ReID$_{RK}$ & 83.3 & 83.7 & 89.9 & 92.7 \\
         P-Duke & 84.5 & 91.8 & 95.2 & 96.2 \\
         P-Duke$_{RK}$ & 88.6 & 89.5 & 92.6 & 93.5 \\
         \hline
         MSMT17 & 76.7 & 90.1 & 94.8 & 95.9 \\
         MSMT17$_{RK}$ & 86.6 & 91.1 & 94.4 & 95.1 \\
         Market-1501 & 90.8 & 95.8 & 98.6 & 99.0 \\
         Market-1501$_{RK}$ & 93.3 & 95.3 & 97.8 & 98.2 \\
         DukeMTMC & 83.3 & 91.0 & 95.7 & 97.1 \\
         DukeMTMC$_{RK}$ & 88.9 & 91.7 & 95.4 & 96.3 \\
         \hline
    \end{tabular}
    \caption{Re-ranking results comparison.}
    \label{tab:rerank_result}
    \end{center}
\end{table}

\section{Attribute prompt template}

In this section, we will list in detail the different template contents mentioned in 4.3.2.

\begin{itemize}
    \item[$\bullet$] Default: ``A photo of a X X X X person.''
    \item[$\bullet$] Longer: ``A person wearing X X X X X X X X X X X X X X X X.''
    \item[$\bullet$] Random: ``X '' * 16
    \item[$\bullet$] Attribute a: ``A person, gender is X, hair is X X, with X X hat, carry X backpack, carry X handbag, wearing X X, X X and X X''
    \item[$\bullet$] Attribute b: ``A X person with X hair, X hat, carry X backpack and X handbag, wearing X X, X X and X X''
    \item[$\bullet$] Attribute c: ``A person, gender is X, hair is X X, with X X hat, carry X backpack, carry X handbag, wearing X X, X X and X X, may be occluded or not''
    \item[$\bullet$] Attribute d: ``A X person with X hair, X hat, carry X backpack and X handbag, wearing X X, X X and X X, may be occluded or not''
\end{itemize}

\section{Model Complexity and Computational Analysis}

\subsection{Parameter Analysis}
During training, the AG-ReID framework introduces an additional attribute encoder module that adds approximately 2.76M parameters. This calculation is based on the following architecture:

The attribute encoder consists of 4 self-attention layers with the following specifications:
\begin{itemize}
    \item Feature dimension: 768 (from ViT-B/16)
    \item Text dimension: 512 (CLIP text encoder output)
    \item Number of attribute class tokens: 11
    \item Transformer layers: 4 layers with multi-head attention
\end{itemize}

The parameter calculation is approximately:
\begin{equation}
\text{Params} = 4 \times (W_{attn} + W_{ffn} + W_{norm})
\end{equation}

where $W_{attn}$ denotes attention weights, $W_{ffn}$ represents feedforward weights, and $W_{norm}$ indicates layer normalization parameters.

This results in approximately 2.76M additional parameters, representing only about 3.2\% of the ViT-B/16 backbone (86M parameters).

\subsection{Inference Efficiency}
Crucially, during inference, AG-ReID only requires the image encoder (ViT-B/16), resulting in:
\begin{itemize}
    \item \textbf{Model size}: Identical to CLIP-ReID (86M parameters)
    \item \textbf{Computational complexity}: Same as other ViT-B/16 based methods
    \item \textbf{Memory usage}: No additional memory overhead during deployment
    \item \textbf{Inference speed}: Equivalent to CLIP-ReID and TransReID
\end{itemize}

The attribute encoder is exclusively used during training and does not affect deployment efficiency, making AG-ReID practically viable for real-world applications without computational overhead.

\section{Ablation Study on $\beta$ Parameter}

We conducted a comprehensive ablation study on the $\beta$ parameter in the loss function $L_{\text{guide}} = L_{\text{id}} + L_{\text{tri}} + L_{\text{CE}} + \beta L_{\text{attrG}}$ to determine its optimal value. The results are presented in Table \ref{tab:beta_ablation}.

\begin{table}[htbp]
    \begin{center}
    \begin{tabular}{c|cccc}
         \hline
         $\beta$ & mAP & R@1 & R@5 & R@10 \\
         \hline
         0.001 & 61.3 & 68.6 & 81.0 & 85.4 \\
         0.002 & 61.4 & 68.6 & 81.2 & 85.7 \\
         0.005 & 61.3 & 68.2 & 81.0 & 85.1 \\
         0.01 & \textbf{63.2} & \textbf{70.4} & \textbf{82.4} & \textbf{86.4} \\
         0.02 & 61.1 & 68.2 & 81.0 & 85.4 \\
         0.05 & 61.5 & 68.4 & 81.7 & 85.7 \\
         0.1 & 61.5 & 68.9 & 81.3 & 85.7 \\
         \hline
    \end{tabular}
    \caption{Ablation study results for different $\beta$ values on Occ-Duke dataset.}
    \label{tab:beta_ablation}
    \end{center}
\end{table}

The experimental results demonstrate that $\beta = 0.01$ achieves the best performance across all evaluation metrics (mAP, Rank-1, Rank-5, and Rank-10). When $\beta$ is too small (0.001, 0.005), the attribute guidance signal is insufficient to effectively guide the image feature learning. Conversely, when $\beta$ is too large (0.02, 0.05), the attribute loss may dominate the training process, potentially interfering with the primary ReID objectives.

The optimal value of $\beta = 0.01$ strikes the right balance between incorporating fine-grained attribute information and maintaining the effectiveness of standard ReID losses, confirming our design choice in the main experiments.